\documentclass[lettersize,journal, twoside]{IEEEtran}
\usepackage{amsmath,amsfonts}
\usepackage{algorithmic}
\usepackage{algorithm}
\usepackage{array}
\usepackage[caption=false,font=normalsize,labelfont=sf,textfont=sf]{subfig}
\usepackage{textcomp}
\usepackage{stfloats}
\usepackage{url}
\usepackage{verbatim}
\usepackage{threeparttable} 
\usepackage{booktabs}       

\usepackage{multirow}       
\usepackage{rotating}       

\usepackage{array}          
\usepackage{graphicx}
\usepackage{cite}
\usepackage{booktabs}
\usepackage{bigstrut,multirow,rotating}

\begin{document}

\title{Selective Kalman Filter: When and How\\ to Fuse Multi-Sensor Information\\ to Overcome Degeneracy in SLAM}

\author{Jie Xu, Guanyu Huang, Wenlu Yu, Xuanxuan Zhang, \\ Lijun Zhao, Ruifeng Li, Shenghai Yuan, Lihua Xie

\thanks{Digital Object Identifier (DOI): see top of this page.}}

\markboth{Journal of \LaTeX\ Class Files,~Vol.~xx, No.~x, August~xxxx}
{Xu \MakeLowercase{\textit{et al.}}: Selective Kalman Filter}

\IEEEpubid{0000--0000/00\$00.00~\copyright~ IEEE}

\maketitle

\begin{abstract}

Research trends in SLAM systems are now focusing more on multi-sensor fusion to handle challenging and degenerative environments. However, most existing multi-sensor fusion SLAM methods mainly use all of the data from a range of sensors, a strategy we refer to as the ``all-in" method. This method, while merging the benefits of different sensors, also brings in their weaknesses, lowering the robustness and accuracy and leading to high computational demands. To address this, we propose a new fusion approach – Selective Kalman Filter – to carefully choose and fuse information from multiple sensors (using LiDAR and visual observations as examples in this paper). For deciding when to fuse data, we implement degeneracy detection in LiDAR SLAM, incorporating visual measurements only when LiDAR SLAM exhibits degeneracy. Regarding degeneracy detection, we propose an elegant yet straightforward approach to determine the degeneracy of LiDAR SLAM and to identify the specific degenerative direction. This method fully considers the coupled relationship between rotational and translational constraints. In terms of how to fuse data, we use visual measurements only to update the specific degenerative states. As a result, our proposed method improves upon the ``all-in" method by greatly enhancing real-time performance due to less processing visual data, and it introduces fewer errors from visual measurements. Experiments demonstrate that our method for degeneracy detection and fusion, in addressing degeneracy issues, exhibits higher precision and robustness compared to other state-of-the-art methods, and offers enhanced real-time performance relative to the ``all-in" method. The code is openly available.
\end{abstract}


\begin{figure}[t]
    \centering
    \includegraphics[width=8.6cm]{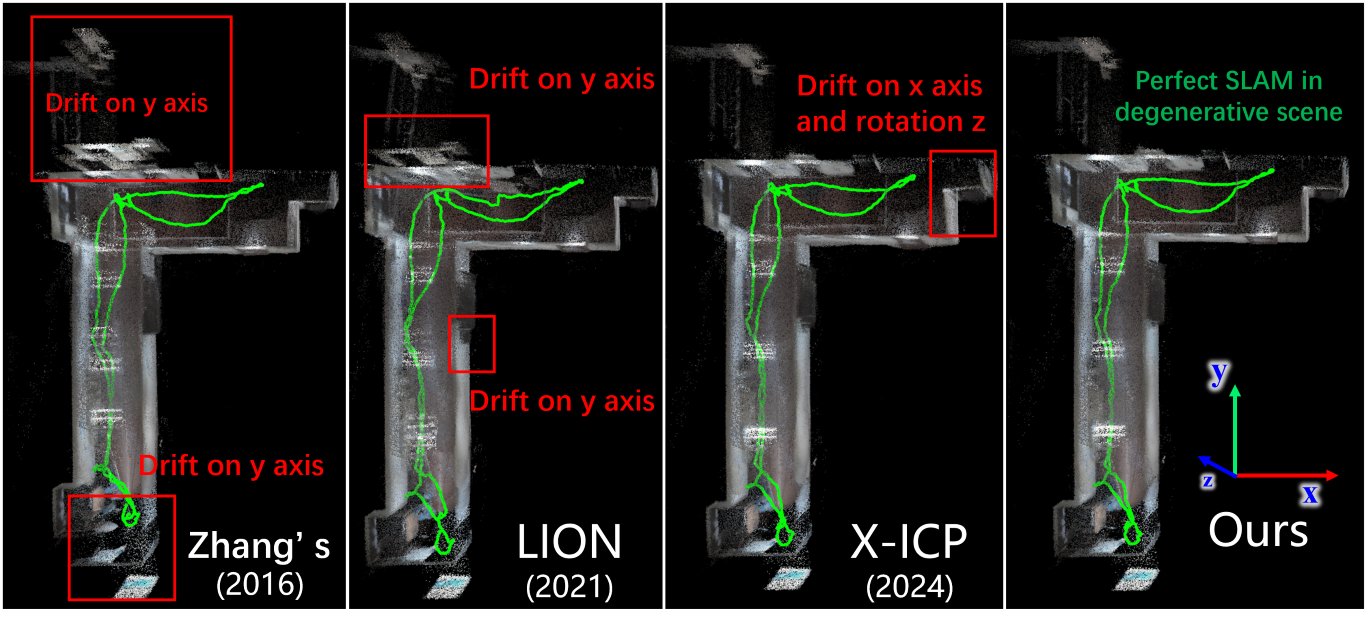}
    \caption{Performance on localization and mapping after applying four different degeneracy detection methods.}
    \label{cover}
\end{figure}
\section{INTRODUCTION}

Simultaneous localization and mapping (SLAM) plays a significant role in robotics, autonomous driving, and Augmented Reality. To address challenging environments and to enhance localization robustness and environmental perception, there is an increasing focus on SLAM systems that fuse multiple sensors \cite{xu2022review}. These systems typically integrate a combination of sensors such as LiDAR, vision, RGB-D camera, and IMU, exemplified by R$^3$LIVE \cite{lin2022r}, FAST-LIVO \cite{zheng2022fast}, LVIO-Fusion \cite{10452777}, RIVER \cite{10404014}, etc.
\IEEEpubidadjcol

Existing Kalman filter research \cite{sasiadek2000sensor,rigatos2007extended} has focused on multi-sensor information fusion, which propose using an Extended Kalman Filter (EKF) and an adaptive fuzzy logic system to fuse odometry and sonar signals. The core idea of these studies are to adaptively adjust the gain. Sun et al. \cite{sun2004multi} present a new multi-sensor optimal information fusion criterion weighted by matrices within the framework of linear minimum variance. By determining the cross-covariance, they then establish the optimal fusion matrix weights. To address the issue of occasional GPS signal loss or inaccuracies, Farhad \cite{aghili2011driftless} applies the KF to incorporate GPS measurements into the data fusion process, utilizing GPS data heavily only when it becomes reliably available. However, these works primarily focus on how to utilize and merge information from multiple sensors, without considering the necessity of fusion or integrating all available data. When fusing high-precision data with moderate-precision data in non-degenerative conditions, the accuracy improvement is often minimal or even negative, and it also incurs additional computational costs. We refer to this indiscriminate fusion of all measurement data as the ``all-in" method. This approach faces two main issues: 

First, the ``all-in" method, by processing a volume of information than single-sensor SLAM, often results in significant computational burden, impacting real-time performance.

Second, while integrating the advantages of various sensors, the ``all-in" method also incorporates their disadvantages. For instance, visual SLAM is frequently affected by motion blur, drastic lighting changes, and dynamic objects, leading to substantial errors or even system failure.

Accordingly, this paper addresses the two aforementioned issues by discussing when and how to selectively fuse multi-sensor information, using a LiDAR and vision-integrated SLAM system as an example. A critical consideration in multi-sensor fusion for SLAM is that individual sensors are prone to localization degeneracy, whereas incorporating different sensor types can provide additional constraints and prevent SLAM degeneracy.

For the ``when" aspect, our solution is to integrate visual information only when the more accurate and robust LiDAR SLAM subsystem shows signs of degeneracy, relying solely on the LiDAR subsystem for pose estimation otherwise.

Regarding ``how" to fuse, if all information is merged, the lower precision of the visual subsystem might reduce the overall accuracy of the LiDAR system. Therefore, our solution is to fuse visual information only in the directions where the LiDAR subsystem is degenerative.

The solution to ``when" requires accurately determining whether degeneracy has occurred in LiDAR SLAM. Existing research in this area includes Zhang's work \cite{zhang2016degeneracy}, which assesses degeneracy by evaluating the condition number of the approximate Hessian matrix. However, due to the differing units and magnitudes of rotation and translation within the state, the eigenvalues of the approximate Hessian matrix couple rotation and translation. Consequently, the derived condition number is not dimensionless, making it challenging to accurately assess degeneracy in practical applications. Similarly, Hinduja's approach \cite{8968577} relies on comparing eigenvalues with thresholds for degeneracy detection, but this method also suffers from the issue of non-uniform dimensions in eigenvalues, complicating the establishment of fixed thresholds. To tackle these challenges, LION \cite{tagliabue2021lion} and Switch-SLAM \cite{10582434} introduce strategies that leverage submatrices of the Hessian matrix for independent degeneracy assessment in rotation and translation. The uniform dimensions within each submatrix endow the eigenvalues with a clear geometric significance, facilitating the application of fixed thresholds for degeneracy analysis. Based on this approach, X-ICP \cite{10328716} also applies normalization and removes mostly noisy data with little contribution. However, these approaches do not account for the high degree of coupling between rotation and translation in point-plane constraints, leading to inaccuracies in determining the extent and direction of degeneracy. MM-LINS \cite{10557776} is our previously proposed work, which includes degeneracy detection, but it does not have a dedicated analysis and experiments specifically for degeneracy detection. Julian \cite{9982257} proposed a learning-based method for degeneracy detection. However, this approach is notably time-consuming and relies solely on current point clouds, which are insufficient for comprehensive degeneracy assessment without incorporating historical point cloud data. Moreover, this method suffers from a lack of generalizability.

Our proposed solution to ``how" entails accurately determining the direction of degeneracy in LiDAR SLAM, incorporating visual measurements exclusively in these degenerative directions. Conversely, the fusion approach by LION \cite{tagliabue2021lion} ceases the use of LiDAR odometry upon any LiDAR degeneracy, switching instead to other odometries. This method is somewhat rudimentary, as relying on lower accuracy and robustness sensor information in non-degenerative directions can actually diminish system performance. The methodologies of Zhang \cite{zhang2016degeneracy}, Hinduja \cite{8968577}, and X-ICP \cite{10328716} advocate for fusing information from other sensors only in degenerative directions, a strategy we consider to be optimal. However, these methods, which are optimization-based, do not address a filtering-based framework, nor can they precisely identify the specific directions of degeneracy. This can result in the true degenerative directions not being adequately constrained following by other sensor measurements.

Therefore, this paper introduces the Selective Kalman Filter to concretely implement the ``when" and ``how" solutions in selective data fusion. ``Selective" here refers to the careful selection of the appropriate timing and data for fusion. We employ the advanced vision-LiDAR fusion scheme R$^3$LIVE as a baseline for deploying the Selective Kalman Filter algorithm. The comparative results of the experiment are shown in Fig. \ref{cover}. The specific contributions of this paper are:
\begin{itemize}{}
\item We introduce a novel multi-sensor fusion approach, named the Selective Kalman Filter, designed to address when and how to selectively fuse data. This method integrates necessary visual information only when LiDAR data is degenerate, significantly enhancing real-time performance while ensuring robustness and accuracy.
\item We introduce an innovative and efficient method for degeneracy detection in LiDAR odometry, which also enables the determination of the degenerative direction. This approach fully considers the coupled relationship between rotational and translational constraints.
\item Through experiments, we demonstrate the accuracy of the proposed degeneration detection method and the efficiency of the Selective Kalman Filter.
\item We open-source the code for broader accessibility and community engagement.
\end{itemize}

\section{NOTATION}
In this study, we simply define the state
\begin{equation}
\mathbf{x}=[\mathbf{R}, \mathbf{t}] \in \mathbf{S O}(3) \times \mathbb{R}^3,
\end{equation}
where $\mathbf{R}$ and $\mathbf{t}$ represent rotation and translation, respectively. Within the framework of Kalman Filter, we define the measurement equation for point-to-plane constraints derived from LiDAR \cite{xu2021fast}, following its linearization, as
\begin{equation}
    z=H\mathbf{x}+v,
\label{origin lidar}
\end{equation}
where $H$ is the Jacobin matrix of dimensions $m \times 6$, with $m$ representing the number of point-to-plane measurements. $z$ is the measurement vector, and $v \in \mathcal{N}\left(\mathbf{0}, R\right)$ comes from the raw LiDAR measurements.

Similarly, the linearized measurement equation for visual data is 
\begin{equation}
b=J\mathbf{x}+w,
\label{origin visual}
\end{equation}
where $J$ is the Jacobin matrix of dimensions $n \times 6$, and $n$ denotes the number of visual measurements. $b$ is the measurement vector, and $w \in \mathcal{N}\left(\mathbf{0}, Q\right)$ comes from the raw visual measurements. In the following sections, we adopt the convention of using uppercase letters to denote matrices and lowercase letters for vectors.

\section{SYSTEM OVERVIEW}
\begin{figure}[ht]
    \centering
    \includegraphics[width=8.2cm]{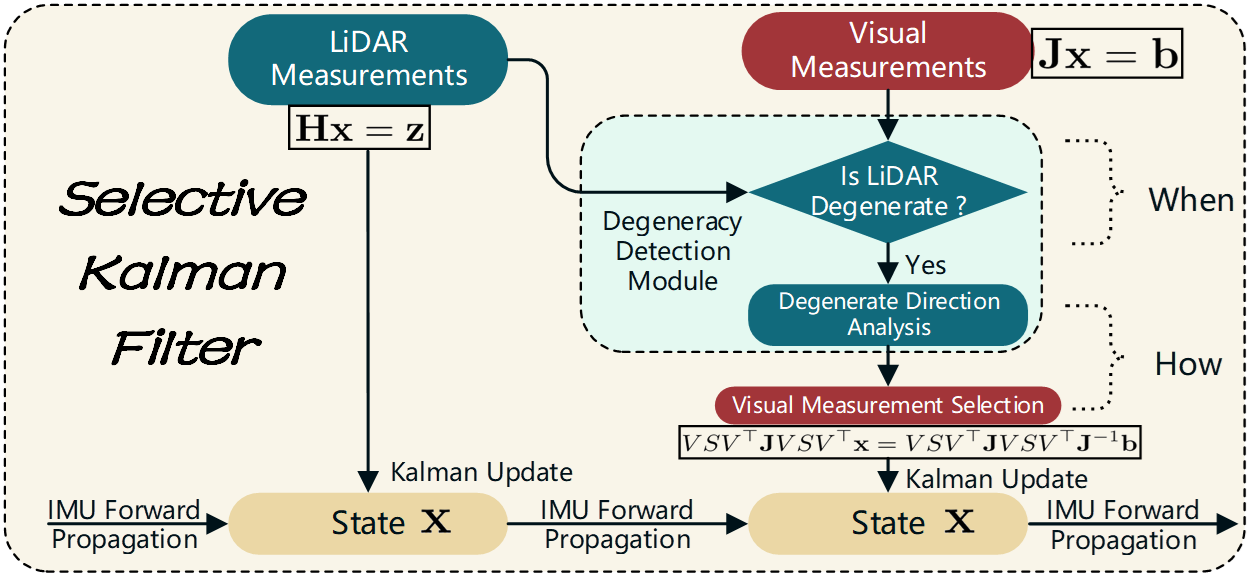}
    \caption{Pipeline of the Selective Kalman Filter in the LIVO system: focusing on ``when" and ``how" to fuse information and LiDAR SLAM degeneracy detection. Teal blocks represent the processing of LiDAR information, while red blocks indicate the handling of visual information.}
    \label{pipeline}
\end{figure}
First, we introduce the ``all-in" method for LiDAR-inertial-vusial SLAM, as exemplified in methodologies like FAST-LIVO \cite{zheng2022fast}, R$^3$LIVE  \cite{lin2022r}. In the absence of visual or LiDAR measurements, similar to FAST-LIO2 \cite{xu2022fast}, the system uses the IMU's acceleration and angular velocity for integration to propagate the state, serving as the prediction step in Kalman Filter. When LiDAR or visual measurements come, the system proceeds with data fusion updates in the Kalman Filter.

To the best of our knowledge, under conditions where neither system degenerates,  LiDAR-inertial odometry (LIO) systems demonstrate higher accuracy and greater robustness compared to visual-inertial odometry (VIO) SLAM \cite{9817108, 9502143}. In integrating the Selective Kalman Filter into the LiDAR-inertial-visual odometry (LIVO) system, we prioritize the LIO subsystem as the core and the VIO subsystem as supplementary. The system's workflow is illustrated in Fig. \ref{pipeline}. Unlike the ``all-in" method, our approach performs degeneracy detection and analysis based on LiDAR measurements:

If LIO subsystem is not degenerative, we opt not to fuse visual measurements. This approach helps prevent the introduction of errors caused by visual issues such as motion blur, overexposure, and incorrect feature matching. Additionally, it reduces computational load, thereby enhancing real-time performance.

If LIO subsystem is degenerative, to minimally introduce visual information and prevent deterioration in state estimation, we analyze the LiDAR's degenerative direction. We then selectively fuse visual data pertinent to this degenerative direction and discard data from non-degenerative directions, which helps prevent a decline in the accuracy of the LiDAR's non-degenerative dimensions.

We refer to this method of selectively updating information fusion as the ``Selective Kalman Filter".

\section{DEGENERACY DETECTION IN LIDAR SLAM}
First, we transform the LiDAR measurement Equation (\ref{origin lidar}) using a weighted linear least squares method, resulting in
\begin{equation}
	H^\top R^{-1}H\hat{\mathbf{x}} = H^\top R^{-1}z,
 \label{weighted lidar measurements}
\end{equation}
where $\hat{\mathbf{x}}$ is the estimated value of state $\mathbf{x}$. For simplification, we have made the following definitions:
\begin{equation}
	\mathbf{H}_\mathbf{I} = H^{\top}R^{-1}H,   \:\mathbf{z}=H^{\top}R^{-1}z.
\label{define1}
\end{equation}
So Equation (\ref{define1}) can be simplified to
\begin{equation}
	\mathbf{H}_\mathbf{I}\hat{\mathbf{x}}=\mathbf{z}.
\end{equation}
Due to noise in real-world scenarios, the matrix $\mathbf{H}_\mathbf{I}$ is invariably a full rank matrix. Consequently, it follows that $\mathbf{H}_\mathbf{I}$ is a $6 \times 6$ real symmetric positive-definite matrix.

Hinduja’s method \cite{8968577} for degenerative detection involves comparing the smallest eigenvalue of $\mathbf{H}_\mathbf{I}$ with a predefined threshold. However, it is important to note that, since the first three dimensions of $\mathbf{H}_\mathbf{I}$ represent rotation and the last three represent translation, the eigenvalues obtained from the eigenvalue decomposition of the $\mathbf{H}_\mathbf{I}$ matrix represent values coupled with both rotation and translation. These values lack a clear physical meaning, as their units are related to both rotation and translation. Furthermore, even within the same eigenvalue decomposition of $\mathbf{H}_\mathbf{I}$, the eigenvalues have different units. Therefore, it becomes challenging to set a reasonable threshold for degeneracy detection. Moreover, the eigenvectors corresponding to these eigenvalues do not represent any actual directions in the real world. Consequently, this method is often impractical for deployment.

To address this issue, LION \cite{tagliabue2021lion} proposes that degeneracy and its direction can be determined by analyzing the diagonal elements of the $\mathbf{H}_\mathbf{I}$. Let us first express the $\mathbf{H}_\mathbf{I}$ matrix in the form of a block matrix
\begin{equation}
	\mathbf{H}_\mathbf{I}=	
	\left[
	\begin{array}{c}
		\mathbf{H}_{rr} \quad \mathbf{H}_{rt}  \\
		\mathbf{H}_{tr} \quad \mathbf{H}_{tt} 
	\end{array}
	\right].
\end{equation}

\begin{figure}[ht]
    \centering
    \includegraphics[width=8.4cm]{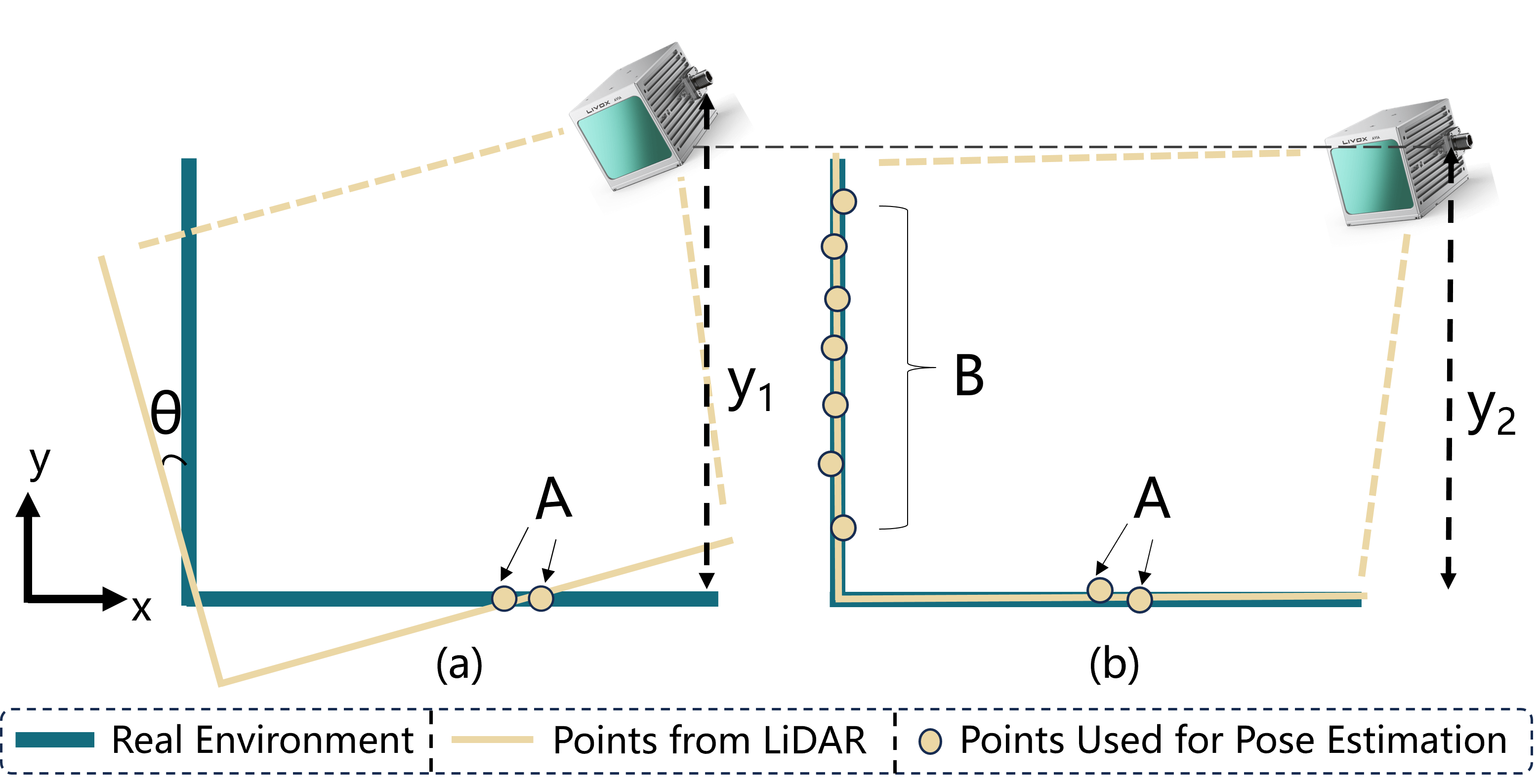}
    \caption{Two-dimensional cross-section diagram illustrating LiDAR $y$-coordinate pose estimation with point clouds: (a) Estimation using only point A, resulting in inaccurate outcomes; (b) Estimation incorporating point A with additional points B, leading to more accurate esitmation.}
    \label{couple}
\end{figure}

LION separately performs eigenvalue decomposition on the $\mathbf{H}_{rr}$ and $\mathbf{H}_{tt}$:
\begin{align}
	\mathbf{H}_{rr} &= {P}_r D_r {P}_r^{\top} &= &{P}_r&\left[\begin{array}{ccc}
		D_{r1} & 0 & 0 \\
		0 & D_{r2}  & 0 \\
		0 & 0  & D_{r3}
	\end{array}\right]{P}_r^{\top}, \\
	\mathbf{H}_{tt} &= {P}_t D_t {P}_t^{\top} &= &{P}_t&\left[\begin{array}{ccc}
		D_{t1} & 0 & 0 \\
		0 & D_{t2}  & 0 \\
		0 & 0  & D_{t3}
	\end{array}\hspace{0.1cm}\right]{P}_t^{\top}.
\end{align}
It then sets two thresholds to assess whether there is degeneration in rotation and translation. It's worth mentioning that this method effectively resolves the issue in Hinduja’s approach where threshold values lack physical significance and cannot be fixed. However, by overlooking the coupled terms $\mathbf{H}_{rt}$ and $\mathbf{H}_{tr}$, it neglects the mutual influence between rotational and translational constraints. Consequently, the degenerative level and direction estimated by this method are often inaccurate in most scenarios. As illustrated in Fig. \ref{couple} (a), according to LION's perspective, the estimation in the $y$-direction is solely dependent on the Jacobians of point-plane residuals in the $y$-direction involving points A, leading to poor estimation in the $y$-direction, as shown by $y_1$. However, in reality, point-plane residuals in the $y$-direction involving points B can improve the estimation of rotation $\theta$ and consequently enhance the $y$-estimation, as indicated by $y_2$. It can be inferred that this latter approach, by considering translation and rotation in a coupled manner, provides higher accuracy for uncertain estimations.

To address the aforementioned issue, we propose a new, elegantly simple, and physically meaningful degeneracy detection method based on covariance information. Initially, we invert the $\mathbf{H}_\mathbf{I}$ matrix, thereby obtaining the covariance matrix $\Sigma$ for state $\mathbf{x}$:
\begin{equation}
	{\Sigma} =	\mathbf{H}_\mathbf{I}^{-1} =
	\left[
	\begin{array}{c}
	{\Sigma}_{rr} \quad {\Sigma}_{rt}  \\
	{\Sigma}_{tr} \quad {\Sigma}_{tt} 
	\end{array}
	\right].
\end{equation}
It should be noted that
\begin{equation}
	{\Sigma}_{rr} =	(\mathbf{H}_{rr} - \mathbf{H}_{rt}\mathbf{H}_{tt}^{-1}\mathbf{H}_{tr})^{-1} \neq \mathbf{H}_{rr}^{-1},
 \label{sigemarr}
\end{equation}
\begin{equation}
	{\Sigma}_{tt} =	(\mathbf{H}_{tt} - \mathbf{H}_{tr}\mathbf{H}_{rr}^{-1}\mathbf{H}_{rt})^{-1} \neq \mathbf{H}_{tt}^{-1}.
  \label{sigematt}
\end{equation}
From Equation (\ref{sigemarr}) and (\ref{sigematt}), it is apparent that our method is not equivalent to LION's approach using a local Hessian matrix. Our method takes into account the coupled relationship between translation and rotation. It is evident from Equations (\ref{sigemarr}) and (\ref{sigematt}) that our method takes into account the coupled relationship between rotation and translation. Additionally, this method has practical physical significance: the diagonal blocks of the covariance matrix represent the variances of the state, while the off-diagonal blocks indicate the correlations between different states. Variance serves as an intuitive measure of degeneracy level. For instance, a high variance indicates significant uncertainty in the estimation of that direction, signifying degeneracy.

We perform eigenvalue decomposition on $\Sigma_{rr}$ and $\Sigma_{tt}$:
\begin{equation}
	\Sigma_{rr}={V}_r \Lambda_r {V}_r^{\top}={V}_r\left[\begin{array}{cccc}
	\lambda_{r1} & 0 & 0 \\
	0 & \lambda_{r2}  & 0 \\
	0 & 0  & \lambda_{r3}
\end{array}\right] {V}_r^{\top},
\end{equation}
\begin{equation}
	\Sigma_{tt}={V}_t \Lambda_t {V}_t^{\top}={V}_t\left[\begin{array}{cccc}
		\lambda_{t1} & 0 & 0 \\
		0 & \lambda_{t2}  & 0 \\
		0 & 0  & \lambda_{t3}
	\end{array}\right] {V}_t^{\top},
\end{equation}
where the units for $\lambda_{ri}\quad(i=1, 2, 3)$ are squared angular units ($\text{rad}^{2})$, while the units for $\lambda_{tj}\quad(j=1, 2, 3)$ are squared length units ($\text{m}^{2}$). 

The columns in $V_r$ and $V_t$ corresponding to $\lambda$ represent respective directions. At this point, both $V_r$ and $V_t$ are rotation matrices with practical physical significance, capable of rotating the original coordinate system to align with these characteristic directions. Consequently, we set two variance thresholds $\theta_r$ and $\theta_t$ based on practical requirements and experience. If $\lambda_{ri} >\theta_r$, it indicates rotational degeneracy, with the corresponding eigenvectors representing the degenerative direction. The same principle applies to translation. It should noted that 
\begin{equation}
	{V}_r \neq {P}_r, {V}_t \neq {P}_t,
 \label{direction}
\end{equation}
which indicates that the principal directions derived from our covariance-based solution differ from those obtained through Hessian-based methods.
 
In summary, this section introduces a degeneracy detection method that has practical physical significance and takes into account coupled constraints. Theoretically, it is capable of more accurately determining the degree and degenerative direction.

\section{SELECTION OF VISUAL MEASUREMENTS}
Initially, we apply a linear weighted least squares approach to the visual measurement Equation (\ref{origin visual}):
\begin{equation}
	J^{\top}Q^{-1}J\hat{\mathbf{x}} = J^{\top}Q^{-1}b.
 \label{visual measurement}
\end{equation}
For simplification, we have made the following definitions:
\begin{equation}
	\mathbf{J}_\mathbf{I} = J^{\top}Q^{-1}J,   \:\mathbf{b}=J^{\top}Q^{-1}b.
 \label{define2}
\end{equation}
So Equation (\ref{visual measurement}) can be simplified to
\begin{equation}
	\mathbf{J}_\mathbf{I}\hat{\mathbf{x}} = \mathbf{b}.
 \label{before}
\end{equation}
Note that $\mathbf{J}_\mathbf{I}$ is always invertible according to the noise in the real world. Therefore, we obtain the estimated state $\hat{\mathbf{x}}$
\begin{equation}
	\hat{\mathbf{x}} = \mathbf{J}_\mathbf{I}^{-1}\mathbf{b}.
\end{equation}
Using the eigenvector matrices of the covariance matrices ${V}_r$ and ${V}_t$, we define a new rotation matrix, denoted as 
\begin{equation}
	{V}=	
	\left[
	\begin{array}{c}
		{V}_r \quad 0  \\
		0 \quad {V}_t 
	\end{array}
	\right].
\end{equation}
It can rotate $\mathbf{x}$'s coordinate system to align with the principal component directions of both translation and rotation:
\begin{equation}
	{V}^{\top}\hat{\mathbf{x}}={V}^{\top}\mathbf{J}_\mathbf{I}^{-1}\mathbf{b}.
\label{new}
\end{equation}
Given the known dimensions corresponding to LiDAR subsystem, we define a selective matrix $S$ to select information in the visual measurement Equation (\ref{new}) that aligns with the degenerative direction, thereby filtering out extraneous data:
\begin{equation}
	{S}{V}^{\top}\hat{\mathbf{x}}={S}{V}^{\top}\mathbf{J}_\mathbf{I}^{-1}\mathbf{b}.
\end{equation}
For instance, if there is degeneracy in one translational and one rotational dimension within the LiDAR subsystem, such as when $\lambda_{r2}>\theta_r $ and $\lambda_{t3}>\theta_t $, we set the selective matrix $S$ as
\begin{equation}
 {S} = \left[\begin{array}{cccccc}
 0 & 0 & 0 & 0 & 0 & 0\\
 0 & 1 & 0 & 0 & 0 & 0\\
 0 & 0 & 0 & 0 & 0 & 0\\
 0 & 0 & 0 & 0 & 0 & 0\\
 0 & 0 & 0 & 0 & 0 & 0\\
 0 & 0 & 0 & 0 & 0 & 1
 \end{array}\right].
\end{equation}
\begin{figure*}[t]
    \centering
    \includegraphics[width=16.2cm]{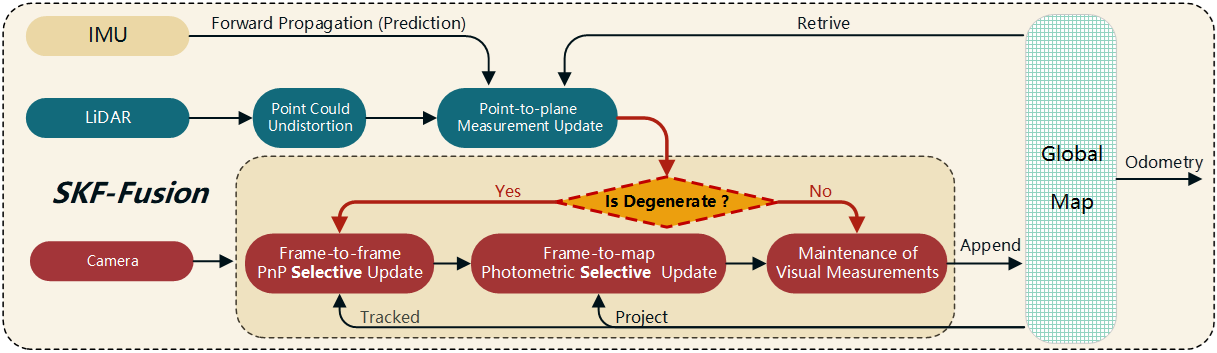}
    \caption{Pipeline of SKF-Fusion.}
    \label{pipeline2}
\end{figure*}
In this matrix, diagonal elements set to 1 correspond to degenerative state dimensions, while those set to 0 correspond to non-degenerative dimensions. We revert the state to its original coordinate system and simultaneously restore the original measurement equation:
\begin{equation}	
\mathbf{J}_\mathbf{I}{V}{S}{V}^{\top}\hat{\mathbf{x}}=\mathbf{J}_\mathbf{I}{V}{S}{V}^{T}\mathbf{J}_\mathbf{I}^{-1}\mathbf{b}.
\end{equation}
To preserve the symmetry of the information matrix, we undertake further processing, resulting in
\begin{equation}
	{V}{S}{V}^{\top}\mathbf{J}_\mathbf{I}{V}{S}{V}^{\top}\hat{\mathbf{x}}={V}{S}{V}^{\top}\mathbf{J}_\mathbf{I}{V}{S}{V}^{\top}\mathbf{J}_\mathbf{I}^{-1}\mathbf{b},
\end{equation}
which can be implified to
\begin{equation}
\mathbf{J}'_\mathbf{I}\hat{\mathbf{x}}=\mathbf{b}',
 \label{after}
\end{equation}
where $\mathbf{J}'_\mathbf{I} = {V}{S}{V}^{\top}\mathbf{J}_\mathbf{I}{V}{S}{V}^{\top}$, $\mathbf{b}' = {V}{S}{V}^{\top}\mathbf{J}_\mathbf{I}{V}{S}{V}^{\top}\mathbf{J}_\mathbf{I}^{-1}\mathbf{b}$.

At this point, we have completed the selection of visual data, retaining only the information pertinent to the direction of LiDAR degeneracy.

\section{SELECTIVE KALMAN FILTER}
First, we describe the conventional form of Kalman Filter. For the visual subsystem, it is assumed that there are pre-existing prior estimates, denoted as $\hat{\mathbf{x}}$ and covariance $P$. Following the methodology used in FAST-LIO \cite{xu2021fast}, the Kalman gain $K$ is calculated as
\begin{equation}
	{K}=({J}^{\top}{Q}^{-1}{J}+{P}^{-1})^{-1}{J}^{\top}{Q}^{-1},
 \label{K}
\end{equation}
leading to the update equation
\begin{equation}
\bar{\mathbf{x}}=\hat{\mathbf{x}}+{K}(b - {J}\hat{\mathbf{x}}).
\label{update}
\end{equation}
By substituting Equation (\ref{K}) into the update Equation (\ref{update}), we obtain
\begin{align}
	\bar{\mathbf{x}} &= \hat{\mathbf{x}} + ({J}^{\top}{Q}^{-1}{J}+{P}^{-1})^{-1}{J}^{\top}{Q}^{-1}b \nonumber \\
	&\quad - ({J}^{\top}{Q}^{-1}{J}+{P}^{-1})^{-1}{J}^{\top}{Q}^{-1} {J}\hat{\mathbf{x}}.
 \label{update2}
\end{align}
According to Equation (\ref{define2}), we can simplify Equation (\ref{update2}) to
\begin{equation}
	\bar{\mathbf{x}} = \hat{\mathbf{x}} + (\mathbf{J}_\mathbf{I}+{P}^{-1})^{-1}\mathbf{b}- (\mathbf{J}_\mathbf{I}+{P}^{-1})^{-1}\mathbf{J}_\mathbf{I}\hat{\mathbf{x}},
\end{equation}
where $\bar{\mathbf{x}}$ is posterior estimated state. Regarding the posterior covariance $\bar{P}$:
\begin{equation}
	\bar{P}=(\mathrm{I}-{K}J)P.
\end{equation}
Substituting in $K$ (Equation (\ref{K})) and simplifying, we obtain
\begin{equation}
	\bar{{P}}=(\mathrm{I}-(\mathbf{J}_\mathbf{I}+{P}^{-1})^{-1}\mathbf{J}_\mathbf{I}){P}.
\end{equation}
By observing the changes in the measurement equation for visual information before (Equation (\ref{before})) and after (Equation (\ref{after})), we can derive the new Selective Kalman Filter formula by simply replacing $\mathbf{J}_\mathbf{I}$ to $\mathbf{J}'_\mathbf{I}$ and $\mathbf{b}$ to $\mathbf{b}'$:
\begin{equation}
	\bar{\mathbf{x}} = \hat{\mathbf{x}} + (\mathbf{J}'_\mathbf{I}+{P}^{-1})^{-1}\mathbf{b}'- (\mathbf{J}'_\mathbf{I}+{P}^{-1})^{-1}\mathbf{J}'_\mathbf{I}\hat{\mathbf{x}},
\end{equation}
\begin{equation}
	\bar{{P}}=(\mathrm{I}-(\mathbf{J}'_\mathbf{I}+{P}^{-1})^{-1}\mathbf{J}'_\mathbf{I}){P}.
\end{equation}

\textit{Remark}: In this paper, matrices subjected to eigenvalue decomposition are exclusively real symmetric matrices, thereby rendering the results of eigenvalue decomposition and singular value decomposition equivalent. Our primary goal is to identify the principal directions of these matrices, which essentially aligns with the physical implications associated with singular values. To streamline the presentation and reduce the number of variables, we uniformly describe the process using eigenvalue decomposition throughout the paper.

\section{LIDAR-INERTIAL-VISUAL ODOMETRY}
Building upon the excellent LIVO framework R$^3$LIVE, we have integrated the Selective Kalman Filter, resulting in the creation of SKF-Fusion, as illustrated in Fig. \ref{pipeline2}. The original R$^3$LIVE framework comprises two parallel subsystems: a direct method for LiDAR-Inertial Odometry (LIO) and a direct method for Visual-Inertial Odometry (VIO). Aspects identical to R$^3$LIVE are not reiterated in this document. Based on this foundation, we conduct degeneracy assessments for the LIO subsystem. If no degeneration is detected, we solely utilize visual information for coloring, while concurrently preserving the map's color information for use during degenerative states. In cases of degeneration, sequential updates between frames and between frames and the map are performed, focusing exclusively on the degenerative state components. In subsequent experiments, we evaluate SKF-Fusion using the R$^3$LIVE dataset.

\begin{figure}[t]
    \centering
    \includegraphics[width=8.4cm]{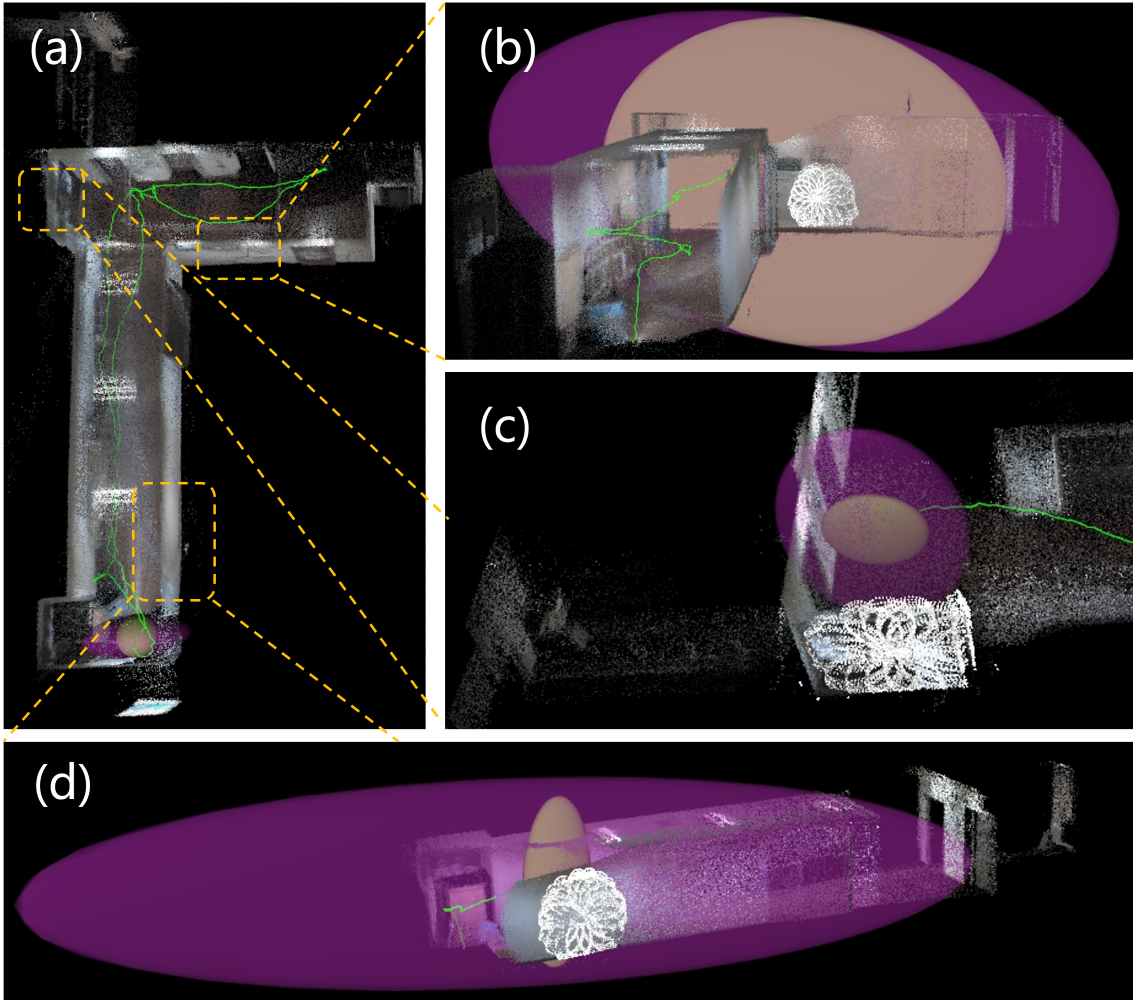}
    \caption{Localization and mapping of SKF-Fusion in the challenging indoor \textit{degenerate\_seq\_02} dataset with evident degeneracy. White point clouds represent the current frame of LiDAR data, purple ellipsoids depict the translational uncertainty representation of our proposed method, and brown ellipsoids depict the translational uncertainty representation of Hessian-based methods (LION, X-ICP). Larger ellipsoids indicate greater uncertainty and more degenerative, and the principal direction of the ellipsoids represents the primary direction of degeneracy. (a) Overall top-down view; (b) and (c) show the uncertainty when the LiDAR faces a wall, and the current frame's point cloud features only a planar characteristic; (d) When facing a wall and the ground, our method identifies the horizontal direction as the most degenerative direction, whereas Hessian-based methods consider it to be the vertical direction.}
    \label{exp01}
\end{figure}
\section{EXPERIMENTS}
Due to the limited availability of datasets containing LiDAR, vision, and IMU information compatible with R$^3$LIVE, this study exclusively utilizes datasets \cite{lin2022r,Lin2023R3LIVEDataset,10669796} provided by R$^3$LIVE for experimental analysis. The handheld device used to record the datasets includes a FLIR Blackfly BFS-u3-13y3c global shutter camera, Livox AVIA LiDAR, and an integrated IMU. Experiments were conducted on a computer equipped with an AMD 7840H CPU and running Ubuntu 20.04. We divided the experiments into two groups: degenerative datasets and normal datasets (the latter may also contain instances of weak degeneracy). For the degenerative datasets, we primarily assess the accuracy of our degeneracy detection method compared to other existing degeneracy detection methods, by measuring the impact on the precision and mapping quality of SKF-Fusion applied SLAM; for the non-degenerative datasets, we evaluate the improvements in accuracy and real-time performance of our SKF-Fusion compared to R$^3$LIVE. The same configuration parameters, including voxel resolution and degeneracy threshold, were used throughout the experiments.
\begin{figure}[t]
    \centering
    \includegraphics[width=8.6cm]{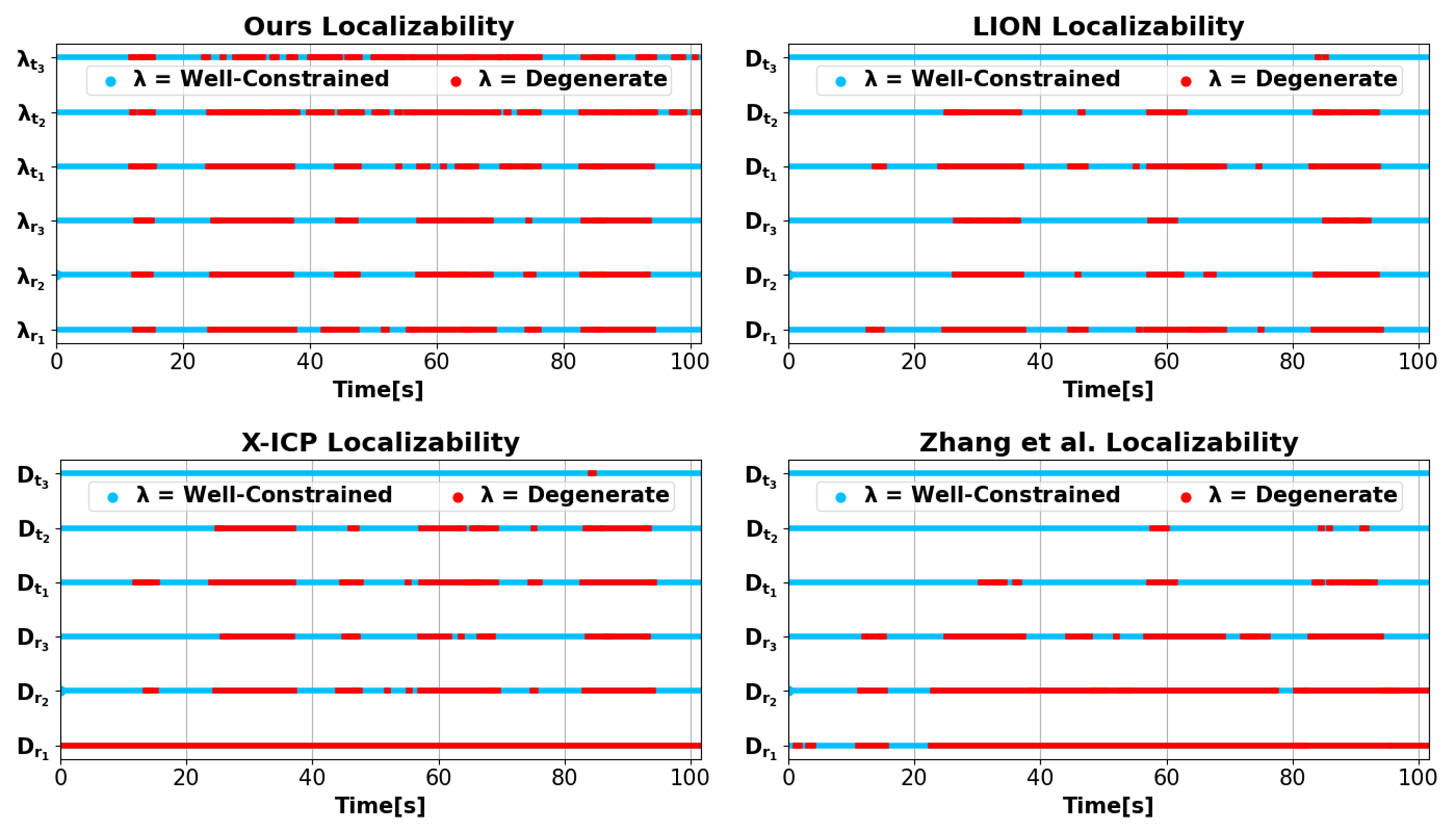}
    \caption{Degeneracy assessment of six degrees of freedom in the \textit{degenerate\_seq\_01} dataset using four different degeneracy detection methods.}
    \label{localizabilitiy}
\end{figure}
\subsection{Comparison in degenerative datasets}
We conducted experiments on degenerative datasets in two outdoor scenes and one indoor scene, named \textit{degenerate\_seq\_00} to \textit{02}. We tested our method, X-ICP, LION, and Zhang's method. If the moments of degeneracy are detected accurately, and the degenerative direction is precisely identified, we can successfully resist degeneracy by integrating visual information. We embedded these degeneracy detection methods within the SKF-Fusion framework, and to ensure a fair comparison, we used the same threshold for all experiments.

\textit{Remark}: Since our method is based on the covariance matrix, while the LION, X-ICP, and Zhang's methods are based on the information matrix, the covariance and information matrices are inversely related. Therefore, the thresholds for these methods should be reciprocal to each other. Specifically, for our method, degeneracy is considered when $\lambda_{r2}>\theta_r$ and $\lambda_{t3}>\theta_t$. For the Hessian-based method, degeneracy is considered when $D_{ri}<\frac{1}{\theta_r}$ and $D_{ti}<\frac{1}{\theta_t}$.
\begin{figure}[b]
    \centering
    \includegraphics[width=8.4cm]{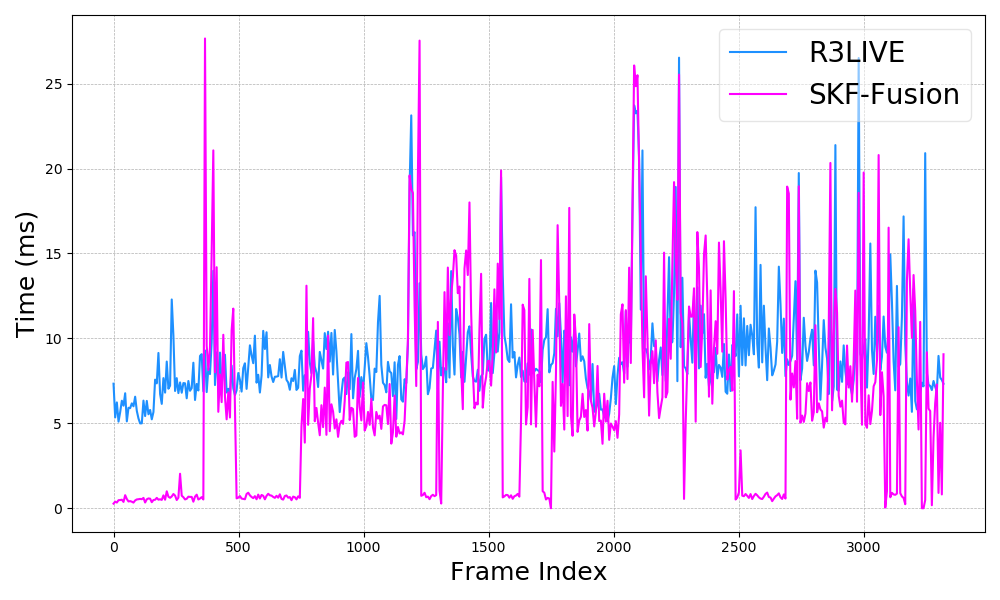}
    \caption{A comparison of the time consumption of the visual subsystems in SKF-Fusion versus R$^3$LIVE within the \textit{degenerate\_seq\_01} dataset.}
    \label{degenerate02time}
\end{figure}
\begin{figure}[t]
    \centering
    \includegraphics[width=8.4cm]{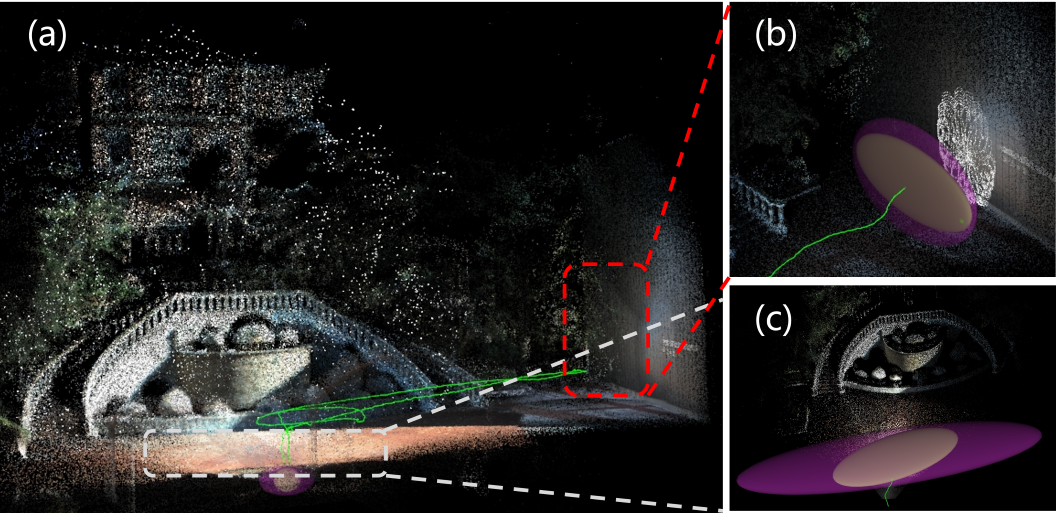}
    \caption{Localization and mapping of SKF-Fusion in the challenging outdoor \textit{degenerate\_seq\_01} dataset. (a) Overview; (b) Representation of uncertainty when facing a wall; (c) Top-Down View. Representation of uncertainty when facing the ground, highlighting the difference in the principal direction of degeneracy estimated by our method compared to Hessian-based methods.}
    \label{exp02}
\end{figure}

In the \textit{degenerate\_seq\_02} dataset, the LiDAR frequently faces a white wall, experiencing significant degeneracy, require visual observations for supplementation. The results of six-degree-of-freedom degeneracy detection using four different methods are illustrated in Fig. \ref{localizabilitiy}, with the localization and mapping outcomes shown in Fig. \ref{cover}. From Fig. \ref{cover}, it is evident that our method successfully resisted degeneracy, achieving high-precision localization and consistent mapping. In contrast, other methods, due to less accurate degeneracy detection, exhibited overlapping artifacts in mapping. In Fig. \ref{localizabilitiy}, compared to our method, LION appears overly optimistic, failing to detect degeneracy when it should. X-ICP, by filtering out many minor noise contributions, detected more instances of degeneracy than LION. However, due to a primary direction that differs from ours (indicated in Equation (\ref{direction})), it still did not achieve optimal results. Zhang's method significantly diverges from other degeneracy detection methods, mistakenly identifying moments of degeneracy as non-degeneracy and vice versa, indicating its relative ineffectiveness in practical degenracy detection tasks.

The translational uncertainty visualization of our covariance-based degeneracy detection method and the Hessian-based methods is shown in the Fig. \ref{exp01}. Corresponding to Equation (\ref{sigemarr}) and (\ref{sigematt}), the Hessian-based methods are more optimistic than ours; hence, the brown ellipsoids are always inside the purple ellipsoids. Moreover, the principal directions of the brown ellipsoids do not align with those of the purple ellipsoids, and there are discrepancies in estimating the direction of greatest degeneracy, especially evident in Fig. \ref{exp01}(d): our method identifies the direction of greatest degeneracy as horizontal, whereas the Hessian-based methods consider it vertical, with the former being evidently more accurate.

In this dataset, our method achieves accuracy comparable to R$^3$LIVE (judged by end-to-end error and mapping results), but because SKF-Fusion only utilizes visual measurements for state updates during LiDAR degeneracy, the time cost of the visual subsystem is significantly reduced. When there is no degeneracy, only a minimal amount of time is spent on visual state maintenance, as shown in Fig. \ref{degenerate02time}.

 \begin{table}[b]
\footnotesize
\centering
\caption{End to End Errors (Meters) in Degenerative Datasets}
\label{tab1}
\begin{threeparttable}
\begin{tabular}{@{}cccc@{}}
\toprule
 & \textit{degenerate\_seq\_00} & \textit{degenerate\_seq\_01} & \textit{degenerate\_seq\_02}   \\ \midrule
R$^3$LIVE        & \underline{0.067 }      & \textbf{0.094}        & \underline{0.100}       \\
Zhang's        & 0.083        & 0.744  & 0.740\\
LION        & 0.072       & 5.781       &      0.116     \\
X-ICP        & 0.082        & 1.214    & 1.827\\
Ours        & \textbf{0.061}       & \underline{0.109}        & \textbf{0.097}      \\
\bottomrule
\end{tabular}
\begin{tablenotes}
\footnotesize
\item[1] The best results are in \textbf{blod}, while the second best are \underline{underlined}.
\end{tablenotes}
\end{threeparttable}
\end{table}

\begin{figure}[t]
    \centering
    \includegraphics[width=8.4cm]{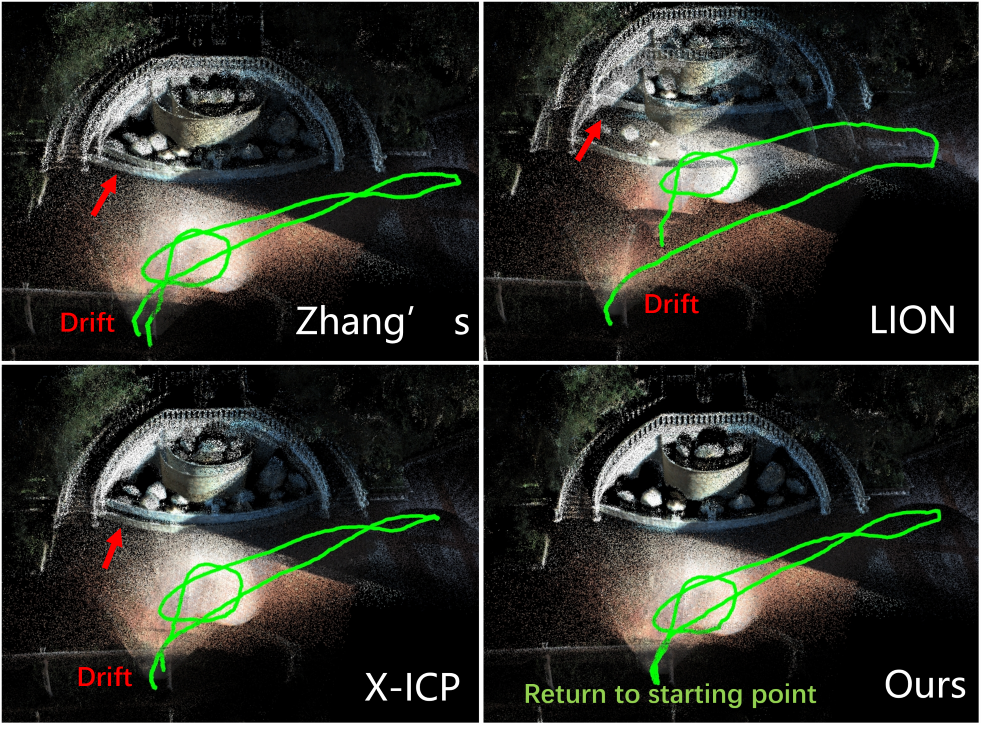}
    \caption{Performance of four different degeneracy detection methods applied in the SKF-Fusion on the \textit{degenerate\_seq\_01} dataset. Except for our method, all other approaches were susceptible to degeneracy, leading to drifts and ultimately failing to return to the starting point.}
    \label{expdata01}
\end{figure}

In the outdoor \textit{degenerate\_seq\_01} dataset, the LiDAR occasionally points towards the ground and walls, resulting in degeneracy across multiple degrees of freedom. The translational uncertainty is depicted as shown in the Fig. \ref{exp02}. We also implemented four different degeneracy detection methods in SKF-Fusion, with the SLAM outcomes presented as indicated in Fig. \ref{expdata01}. It is evident that our method maintains high end-to-end accuracy and generates consistent maps without any ghosting effects.

In these three datasets, we evaluated the accuracy of R$^3$LIVE and four degeneracy detection methods applied within SKF-Fusion. As shown in Table \ref{tab1}, among the SKF-Fusion implementations utilizing the four degeneracy detection methods, the method we proposed achieved the best SLAM accuracy. This close resemblance in accuracy between our method and R$^3$LIVE can be attributed to the datasets not introducing visual disturbances such as drastic lighting changes.

\begin{figure}[b]
    \centering
    \includegraphics[width=8.4cm]{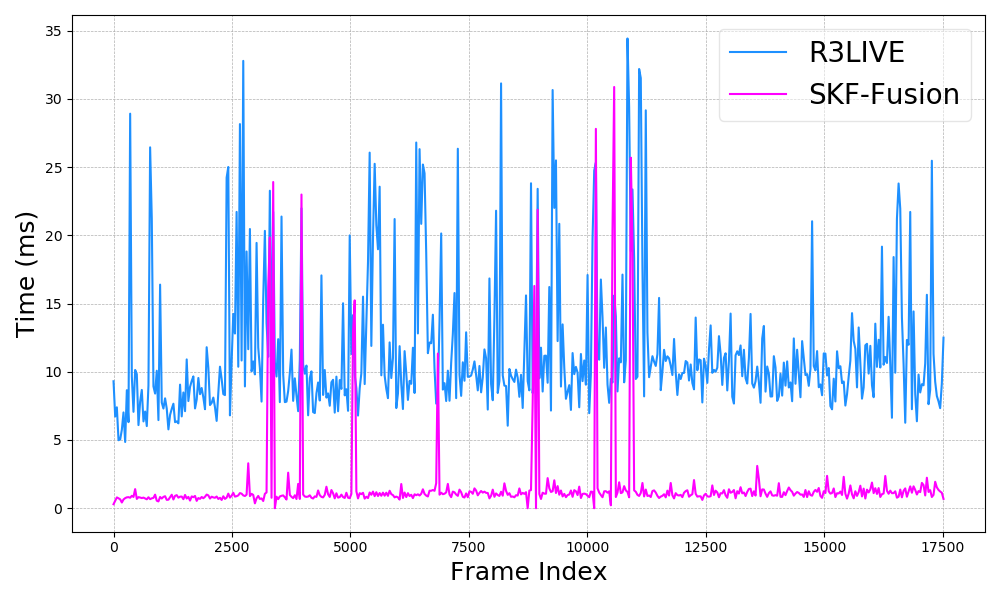}
    \caption{A comparison of the time consumption of the visual subsystems in SKF-Fusion versus R$^3$LIVE within the \textit{hku\_main\_building} dataset.}
    \label{hkubuildingtime}
\end{figure}
\begin{table}[t]
\footnotesize
\centering
\caption{VIO Per-frame Cost Time and End to End Errors}
\label{tab2}
\begin{threeparttable}
\begin{tabular}{@{}ccccc@{}}
\toprule
 & \multicolumn{2}{c}{\textit{SKF-Fusion}} & \multicolumn{2}{c}{\textit{R$^3$LIVE}} \\
 & ms & m & ms & m \\ \midrule
\textit{hku\_main\_building} & {1.071} & \textbf{0.056} & {11.953} & 0.117 \\
\textit{hku\_campus\_seq\_00} & {0.844} & \textbf{0.101} & {9.544} & 0.103 \\
\textit{hku\_campus\_seq\_02} & {0.891} & \textbf{0.109} & {10.691} & 0.12 \\
\textit{hku\_campus\_seq\_03} & {0.844} & \textbf{0.062} & {9.973} & 0.0701 \\
\textit{hku\_park\_00} & {0.797} & 0.08 & {13.113} & 0.076 \\
\textit{hku\_park\_01} & {0.861} & \textbf{0.553} & {11.744} & 0.61 \\
\bottomrule
\end{tabular}
\end{threeparttable}
\end{table}
\subsection{Comparison in normal datasets}
In normal datasets, significant degeneracy rarely occurs. Thus, these datasets are not effective for distinguishing between the performance of various degeneracy detection methods, nor do they significantly differentiate accuracy levels. However, they can be used to assess the real-time performance of R$^3$LIVE compared to SKF-Fusion. As illustrated in Fig. \ref{hkubuildingtime}, in the \textit{hku\_main\_building} dataset (lasting approximately 1750 seconds), the time expenditure of the visual subsystem in SKF-Fusion is considerably less than that of R$^3$LIVE, resulting in substantial computational resource savings. In most scenarios where LIO does not degenerate, the visual subsystem spends only 0-2 ms on visual information maintenance without performing state updates; under degenerative conditions, an additional 5-30 ms is required for integrating visual information with the radar subsystem to update the state. Time consumption and end-to-end error in other datasets are presented in Table \ref{tab2}. From Table \ref{tab2}, it can also be observed that due to predominantly relying on more accurate LiDAR measurements and less on visual measurements in most cases, there is a slight improvement in accuracy.

\section{CONCLUSIONS}
Building on the traditional foundation of ``all-in" fusion SLAM involving LiDAR, vision, and IMU, this study introduces an innovative degeneracy detection module based on covariance information, along with a Selective Kalman Filter for the selective updating of degenerative states. Compared to other existing degeneracy detection methods, empirical evidence demonstrates that our proposed method significantly enhances the accuracy of degeneracy detection, thereby improving both the precision and robustness of localization. Notably, our approach substantially augments the real-time performance of the visual subsystem compared to the ``all-in"  method. It is important to mention that the current threshold selection does not robustly adapt to LIO across different voxel resolutions, highlighting an area for further research.



\section*{DISCUSSION AND FUTURE WORK}
The Selective Kalman Filter extends beyond the fusion of LiDAR and vision, being applicable to a broad array of combinations, such as LiDAR with wheel encoders, Pedestrian Dead Reckoning (PDR), GPS, Ultra-Wideband (UWB), and more. 


\bibliographystyle{IEEEtran}
\bibliography{reference}

\begin{thebibliography}{10}
\providecommand{\url}[1]{#1}
\csname url@samestyle\endcsname
\providecommand{\newblock}{\relax}
\providecommand{\bibinfo}[2]{#2}
\providecommand{\BIBentrySTDinterwordspacing}{\spaceskip=0pt\relax}
\providecommand{\BIBentryALTinterwordstretchfactor}{4}
\providecommand{\BIBentryALTinterwordspacing}{\spaceskip=\fontdimen2\font plus
\BIBentryALTinterwordstretchfactor\fontdimen3\font minus \fontdimen4\font\relax}
\providecommand{\BIBforeignlanguage}[2]{{%
\expandafter\ifx\csname l@#1\endcsname\relax
\typeout{** WARNING: IEEEtran.bst: No hyphenation pattern has been}%
\typeout{** loaded for the language `#1'. Using the pattern for}%
\typeout{** the default language instead.}%
\else
\language=\csname l@#1\endcsname
\fi
#2}}
\providecommand{\BIBdecl}{\relax}
\BIBdecl

\bibitem{xu2022review}
X.~Xu, L.~Zhang, J.~Yang, C.~Cao, W.~Wang, Y.~Ran, Z.~Tan, and M.~Luo, ``A review of multi-sensor fusion slam systems based on 3d lidar,'' \emph{Remote Sensing}, vol.~14, no.~12, p. 2835, 2022.

\bibitem{lin2022r}
J.~Lin and F.~Zhang, ``R 3 live: A robust, real-time, rgb-colored, lidar-inertial-visual tightly-coupled state estimation and mapping package,'' in \emph{2022 International Conference on Robotics and Automation (ICRA)}.\hskip 1em plus 0.5em minus 0.4em\relax IEEE, 2022, pp. 10\,672--10\,678.

\bibitem{zheng2022fast}
C.~Zheng, Q.~Zhu, W.~Xu, X.~Liu, Q.~Guo, and F.~Zhang, ``Fast-livo: Fast and tightly-coupled sparse-direct lidar-inertial-visual odometry,'' in \emph{2022 IEEE/RSJ International Conference on Intelligent Robots and Systems (IROS)}.\hskip 1em plus 0.5em minus 0.4em\relax IEEE, 2022, pp. 4003--4009.

\bibitem{10452777}
H.~Zhang, L.~Du, S.~Bao, J.~Yuan, and S.~Ma, ``Lvio-fusion:tightly-coupled lidar-visual-inertial odometry and mapping in degenerate environments,'' \emph{IEEE Robotics and Automation Letters}, vol.~9, no.~4, pp. 3783--3790, 2024.

\bibitem{10404014}
T.~Wen, Y.~Fang, B.~Lu, X.~Zhang, and C.~Tang, ``Liver: A tightly coupled lidar-inertial-visual state estimator with high robustness for underground environments,'' \emph{IEEE Robotics and Automation Letters}, vol.~9, no.~3, pp. 2399--2406, 2024.

\bibitem{sasiadek2000sensor}
J.~Sasiadek and P.~Hartana, ``Sensor data fusion using kalman filter,'' in \emph{Proceedings of the Third International Conference on Information Fusion}, vol.~2.\hskip 1em plus 0.5em minus 0.4em\relax IEEE, 2000, pp. WED5--19.

\bibitem{rigatos2007extended}
G.~Rigatos and S.~Tzafestas, ``Extended kalman filtering for fuzzy modelling and multi-sensor fusion,'' \emph{Mathematical and computer modelling of dynamical systems}, vol.~13, no.~3, pp. 251--266, 2007.

\bibitem{sun2004multi}
S.-L. Sun and Z.-L. Deng, ``Multi-sensor optimal information fusion kalman filter,'' \emph{Automatica}, vol.~40, no.~6, pp. 1017--1023, 2004.

\bibitem{aghili2011driftless}
F.~Aghili and A.~Salerno, ``Driftless 3-d attitude determination and positioning of mobile robots by integration of imu with two rtk gpss,'' \emph{IEEE/ASME Transactions on Mechatronics}, vol.~18, no.~1, pp. 21--31, 2011.

\bibitem{zhang2016degeneracy}
J.~Zhang, M.~Kaess, and S.~Singh, ``On degeneracy of optimization-based state estimation problems,'' in \emph{2016 IEEE International Conference on Robotics and Automation (ICRA)}.\hskip 1em plus 0.5em minus 0.4em\relax IEEE, 2016, pp. 809--816.

\bibitem{8968577}
A.~Hinduja, B.-J. Ho, and M.~Kaess, ``Degeneracy-aware factors with applications to underwater slam,'' in \emph{2019 IEEE/RSJ International Conference on Intelligent Robots and Systems (IROS)}, 2019, pp. 1293--1299.

\bibitem{tagliabue2021lion}
A.~Tagliabue, J.~Tordesillas, X.~Cai, A.~Santamaria-Navarro, J.~P. How, L.~Carlone, and A.-a. Agha-mohammadi, ``Lion: Lidar-inertial observability-aware navigator for vision-denied environments,'' in \emph{Experimental Robotics: The 17th International Symposium}.\hskip 1em plus 0.5em minus 0.4em\relax Springer, 2021, pp. 380--390.

\bibitem{10582434}
J.~Lee, R.~Komatsu, M.~Shinozaki, T.~Kitajima, H.~Asama, Q.~An, and A.~Yamashita, ``Switch-slam: Switching-based lidar-inertial-visual slam for degenerate environments,'' \emph{IEEE Robotics and Automation Letters}, vol.~9, no.~8, pp. 7270--7277, 2024.

\bibitem{10328716}
T.~Tuna, J.~Nubert, Y.~Nava, S.~Khattak, and M.~Hutter, ``X-icp: Localizability-aware lidar registration for robust localization in extreme environments,'' \emph{IEEE Transactions on Robotics}, vol.~40, pp. 452--471, 2024.

\bibitem{10557776}
Y.~Ma, J.~Xu, S.~Yuan, T.~Zhi, W.~Yu, J.~Zhou, and L.~Xie, ``Mm-lins: a multi-map lidar-inertial system for over-degenerate environments,'' \emph{IEEE Transactions on Intelligent Vehicles}, pp. 1--11, 2024.

\bibitem{9982257}
J.~Nubert, E.~Walther, S.~Khattak, and M.~Hutter, ``Learning-based localizability estimation for robust lidar localization,'' in \emph{2022 IEEE/RSJ International Conference on Intelligent Robots and Systems (IROS)}, 2022, pp. 17--24.

\bibitem{xu2021fast}
W.~Xu and F.~Zhang, ``Fast-lio: A fast, robust lidar-inertial odometry package by tightly-coupled iterated kalman filter,'' \emph{IEEE Robotics and Automation Letters}, vol.~6, no.~2, pp. 3317--3324, 2021.

\bibitem{xu2022fast}
W.~Xu, Y.~Cai, D.~He, J.~Lin, and F.~Zhang, ``Fast-lio2: Fast direct lidar-inertial odometry,'' \emph{IEEE Transactions on Robotics}, vol.~38, no.~4, pp. 2053--2073, 2022.

\bibitem{9817108}
L.~R. Agostinho, N.~M. Ricardo, M.~I. Pereira, A.~Hiolle, and A.~M. Pinto, ``A practical survey on visual odometry for autonomous driving in challenging scenarios and conditions,'' \emph{IEEE Access}, vol.~10, pp. 72\,182--72\,205, 2022.

\bibitem{9502143}
T.-M. Nguyen, M.~Cao, S.~Yuan, Y.~Lyu, T.~H. Nguyen, and L.~Xie, ``Viral-fusion: A visual-inertial-ranging-lidar sensor fusion approach,'' \emph{IEEE Transactions on Robotics}, vol.~38, no.~2, pp. 958--977, 2022.

\bibitem{Lin2023R3LIVEDataset}
Z.~Lin, ``{R3LIVE Dataset},'' \url{https://github.com/ziv-lin/r3live_dataset}, 2023, accessed: 2023-02-27.

\bibitem{10669796}
J.~Lin and F.~Zhang, ``R $^{3}$ live++: A robust, real-time, radiance reconstruction package with a tightly-coupled lidar-inertial-visual state estimator,'' \emph{IEEE Transactions on Pattern Analysis and Machine Intelligence}, pp. 1--18, 2024.

\end{thebibliography}
\end{document}